\pdfoutput=1
\documentclass[10pt,twocolumn,letterpaper]{article}

\usepackage[pagenumbers]{cvpr}

\definecolor{cvprblue}{rgb}{0.21,0.49,0.74}
\usepackage[pagebackref,breaklinks,colorlinks,allcolors=cvprblue]{hyperref}

\usepackage{xcolor}
\usepackage{booktabs}
\usepackage{amsmath}
\usepackage{graphicx}
\newcommand{\ours}{NVA}
\newcommand{\dev}{FLUX.1-dev}
\newcommand{\qwen}{Qwen-Image}
\newcommand{\afk}{A4K}
\newcommand{\up}{$\uparrow$}
\newcommand{\down}{$\downarrow$}

\def\confName{CVPR}
\def\confYear{2027}

\title{Native View Attention: Training-Free Ultra-High-Resolution Image Generation in Linear Time}

\author{
Xuanming Shang\textsuperscript{1} \quad
Weijia Zhang\textsuperscript{1} \quad
Yuehao Liu\textsuperscript{1} \quad
Xiangxuan Ren\textsuperscript{1} \quad
Chao Ma\textsuperscript{1,\ensuremath{\dagger}}\\[3pt]
{\small \textsuperscript{1}MoE Key Lab of Artificial Intelligence, Institute of AI, Shanghai Jiao Tong University, Shanghai, China}\\[2pt]
{\small \texttt{\{sxm2021, chaoma\}@sjtu.edu.cn}}\\[-4pt]
}

\usepackage{placeins}
\begin{document}
\maketitle
{\renewcommand{\thefootnote}{}\footnotetext{\textsuperscript{\ensuremath{\dagger}}Corresponding author.}}
\begin{abstract}
Diffusion transformers sampled beyond their training resolution lose global
structure and produce artifacts, and their attention cost grows quadratically
with the number of tokens. We trace both problems to one cause. Attention
counts every token once, and in training each token stands for a fixed area of
the image. At a higher resolution, the same area is split into many
near-identical tokens, so dense attention counts it many times. Comparing
attention at both resolutions, we find that the prompt, whose tokens do not
multiply, loses most of its attention. Adapting position encodings or
sharpening attention does not restore it. Window attention avoids the cost but
counts nothing outside its window, so content repeats. We propose \emph{Native
View Attention} (NVA), a training-free attention operator that counts every
image area once, as in training. Where detail is not needed, the tokens of an
area are pooled into one. Near each token and in the few most relevant distant
regions, they are kept and share the count of their area. Pooling also
fixes the number of tokens each token attends to, which makes the cost linear.
On FLUX.1-dev and Qwen-Image, NVA surpasses prior training-free methods in
global fidelity, prompt alignment and human preference, and is the fastest, up
to $8.4\times$ faster than dense attention at $8192^2$.
Project page: \url{https://nvattention.github.io/}.
\end{abstract}

\section{Introduction}
\label{sec:intro}

\begin{figure}[t]
\centering
\includegraphics[width=0.94\linewidth]{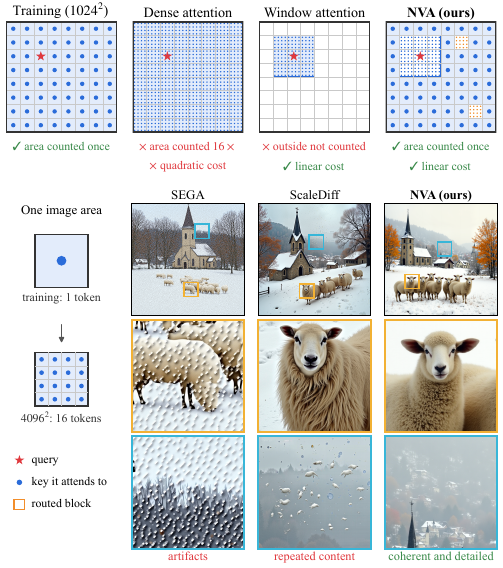}
\caption{What an image query (red star) attends to, and the $4096^2$ image that
results on \dev{}. In training, every image area holds one token and is counted
once. At $4096^2$, the same area holds 16 tokens. Dense attention counts every
area 16 times, window attention does not count the areas outside its window,
and \ours{} counts every area once, with routed blocks in orange. Below each
attention pattern, a method that uses it is shown with two zoomed regions. Red labels name the failure.}
\label{fig:teaser}
\vspace{-8pt}
\end{figure}

Diffusion transformers (DiTs)~\cite{dit} such as \dev{}~\cite{flux} and
\qwen{}~\cite{qwenimage} generate high-fidelity images at their native
resolution of about $1024^2$ pixels, the resolution they are trained at. Applications such as print and large
displays, however, need ultra-high resolutions of $4096^2$ pixels or more. Training at such resolutions is prohibitive, because high-resolution data are scarce and the cost of attention grows quadratically with the number
of tokens. Although models trained or fine-tuned at $4096^2$ have
appeared~\cite{pixartsigma,ultrapixel,diffusion4k,ultraflux}, a more practical alternative is to
sample a pretrained model at the target resolution without training. This
training-free setting, however, faces two problems. The image loses its global
structure and fills with artifacts (Fig.~\ref{fig:teaser}), and a $4096^2$ image
has $16\times$ as many tokens as a $1024^2$ one, so dense attention costs
$256\times$ more and dominates the runtime. We trace both problems to one cause
and solve them together.

The cause lies in how attention counts the image. Each token, as a query,
distributes its attention over a set of keys, its receptive field, through one
softmax in which every key counts once. In training, every key stands for a
fixed area of the image, so every area counts once. At $4096^2$, the same area
is split into 16 tokens with nearly identical content, and dense attention
counts it 16 times (Fig.~\ref{fig:teaser}). The text makes this miscount
measurable, because its tokens do not multiply. On \dev{}, the text attention
share, the fraction of an image query's attention that goes to the text, falls
from about 16\% at the native resolution to 2--4\% (Fig.~\ref{fig:analysis}).
Related effects are known. A growing number of tokens was found to raise the attention
entropy~\cite{jin2023training} and to dilute the attention on the
text~\cite{tide}. These works compensate for the effect and keep dense
attention. We instead locate the cause in the keys themselves, which also
explains why window attention fails and leads to a remedy that lowers the cost.

Existing training-free methods leave this count unchanged. Position
extrapolation~\cite{luminanext,dype,sega,sigma,ultraimage} adapts the position
encoding to the target resolution, and attention
scaling~\cite{jin2023training,imax,ultraimage,sega,tide} sharpens the softmax.
Neither changes the keys, so every area still counts 16 times, the text
attention share stays far below its native value (Fig.~\ref{fig:analysis}) and
the cost stays quadratic. Sparse attention~\cite{spargeattn,xattention,pisa}
lowers the cost but aims to reproduce dense attention, and so inherits its
count. Window attention~\cite{hrdit,scalediff} attends only to a window around
the query, which avoids the quadratic cost. Image areas outside the window,
however, are not counted at all. Each window is then generated like an
independent image, and content repeats across windows
(Fig.~\ref{fig:teaser}). Many of these methods run in a
cascade~\cite{hiflow,imax,hrdit,scalediff}, which first generates an image at
the native resolution and then refines it at the target
resolution. The cascade fixes the layout, but the refinement still uses dense
or window attention, so the details form under a wrong count.

The remedy is to count every image area once, as in training, over the
whole image. We propose \emph{Native View Attention} (\ours{}), a training-free
attention operator that follows this rule in the refinement stage of the same
cascade (Fig.~\ref{fig:teaser}). Most of the image only provides context to a
query. There, \ours{} pools the 16 tokens of each area into one key, which
counts once. This far field gives the query the view of the whole image that it
had in training. Details form near the query, so its near field keeps the
target tokens. A query also attends strongly to a few distant regions, which
its attention in the native-resolution stage reveals. These routed blocks keep
their target tokens as well. The kept tokens share the count of their area. A
routed token counts as one sixteenth of a pooled key, and the near field as a
whole receives the attention that native attention gives to its area. All keys
share one softmax, as in training. The number of keys per query no longer grows
with the resolution, so the attention cost is linear in the number of tokens.

We apply \ours{} to \dev{} and \qwen{} without any training and compare it
with eight training-free methods under a unified protocol. On Aesthetic-Eval@4096, \ours{}
leads the training-free methods on both models in whole-image fidelity (FID,
IS), prompt alignment (CLIP score), human preference (HPS) and perceptual
quality (MUSIQ). On \dev{}, it also exceeds the 4K-fine-tuned Diffusion-4K on
all seven quality metrics and is best on all five metrics of LAION-1K. \ours{}
is also the fastest training-free method. On \dev{}, it generates $4096^2$ and
$8192^2$ images $2.5\times$ and $8.4\times$ faster than the same cascade with
dense attention.
Our contributions are as follows:
\begin{itemize}
\item We trace the quality and cost problems of training-free
ultra-high-resolution generation to one cause. Dense attention counts every
image area many times, and window attention does not count the areas outside
its window.
\item We propose \ours{}, a training-free attention operator that counts
every image area once. It pools the tokens of distant areas and weights the
tokens it keeps by their area, at a cost linear in the number of tokens.
\item Experiments on two DiTs and two benchmarks show that \ours{} surpasses
existing training-free methods in both quality and speed.
\end{itemize}

\section{Related Work}
\label{sec:related}

\paragraph{Training-free high-resolution generation.}
Early work targeted U-Net diffusion models. It traced high-resolution failures to the limited receptive field of convolutions and enlarged it by re-dilating the kernels~\cite{scalecrafter,fouriscale} or by fusing receptive scales~\cite{freescale}. Other methods fuse overlapping patches~\cite{multidiffusion,demofusion}, recently with few-step denoising~\cite{pixelrush}, or rescale the attention temperature with the number of tokens~\cite{jin2023training}. For DiTs, one line of work
adapts the position encoding to the target resolution~\cite{dype,sega,sigma,ultraimage}, following context extension in language models~\cite{positioninterp,yarn,luminanext}, often together with
attention scaling~\cite{imax,ultraimage,sega}. TIDE~\cite{tide} rebalances text and image tokens against attention dilution, and ResDiT~\cite{resdit} adds local attention at the native resolution. Cascades generate a native image first and guide the
high-resolution trajectory with it~\cite{hiflow,imax}, some combined with
window attention~\cite{hrdit,scalediff}. Models fine-tuned at $4096^2$~\cite{diffusion4k,ultraflux} or trained for it~\cite{pixartsigma,ultrapixel}
avoid the problem with additional training. These methods keep dense attention or restrict it to a window, and thus count image areas many times or not at all, whereas \ours{} counts each once.

\paragraph{Efficient and distance-dependent attention.}
Training-free sparse attention skips query--key blocks with low
estimated weight~\cite{spargeattn,draftattention,xattention,sparsevideogen,sparsevideogen2}. Others approximate the skipped blocks instead of dropping them~\cite{rectifiedspaattn,pisa,svgear}. Window attention uses fixed
windows~\cite{natten,sta,swin}, lets the compute density decay with distance~\cite{radialattn}, caches the residual to full
attention~\cite{ditfastattn}, or is obtained by fine-tuning~\cite{clear}. Token
downsampling reduces the keys and values~\cite{todo,pvt}, and linear attention
replaces the softmax in a model trained from scratch~\cite{sana}. Sparse methods aim to match dense attention at a lower cost and window methods restrict it to a window, so neither corrects how image areas are counted beyond the native resolution. Attending to near and far tokens at different
granularities is a classical alternative. Following the fast multipole method,
several transformers compute near-field attention exactly and represent the far
field with low-rank or coarsened terms~\cite{fmmformer,htransformer,fastmultipole},
and focal attention~\cite{focal}, sparse transformers with global tokens~\cite{longformer,bigbird}, native
sparse attention~\cite{nsa} and LLSA~\cite{llsa} pair fine local keys with
coarse or selected distant keys. Token merging~\cite{tome} adds the logarithm of
the merged size to the attention logits, so that a merged token counts as the tokens it replaces and the attention before merging is preserved. Most of these designs are trained together with their
attention pattern. \ours{} applies the principle to a pretrained model without
training, with the opposite purpose. Its weights set the count of every image area to its value in training instead of preserving that of dense attention. The linear cost follows from
restoring the pretrained receptive field.

\section{Method}
\label{sec:method}

\begin{figure}[t]
\centering
\includegraphics[width=0.95\linewidth]{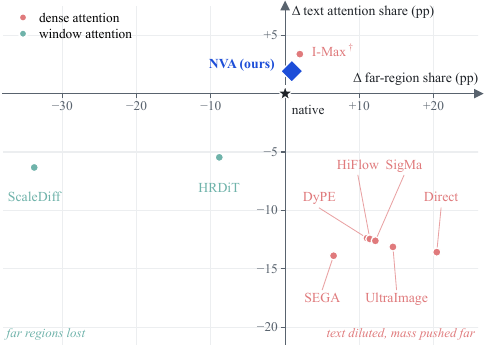}
\caption{Change of attention of an image query at $4096^2$ on \dev{},
relative to native $1024^2$ attention (origin), under the attention operator of
each compared training-free method. The x-axis gives the change
of the attention share on image regions at least 16 native tokens away, and the
y-axis that of the text attention share, in percentage points. $^\dagger$I-Max adds 16 copies of the text
tokens, which offsets the dilution of the text.}
\label{fig:analysis}
\vspace{-8pt}
\end{figure}

\begin{figure*}[t]
\centering
\includegraphics[width=\textwidth]{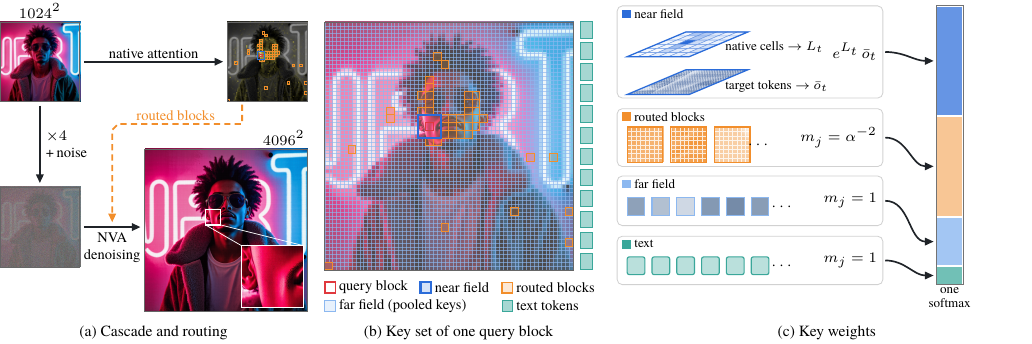}
\caption{\textbf{Overview of \ours{}.} (a) Cascade and routing. The first
stage generates a $1024^2$ image, whose latent is upsampled, re-noised and
denoised at $4096^2$ with \ours{}. The overlay shows the native attention of
one query in one layer and head (log scale), from which its routed blocks are
selected. The inset magnifies the near field in the $4096^2$ output. (b) Key
set of the query block in (a). Some routed blocks lie next to the near field,
and others lie on the other cheek, the lenses, the hair and the clothing.
(c) Key weights. Each key is drawn with the native area it covers, which
sets its weight. The near field takes its total weight $e^{L_t}$ from native
attention over its cells and its inner distribution $\bar o_t$ from the target
tokens. The bar shows how the native attention of this query splits over the
four parts of one softmax.}
\label{fig:method}
\vspace{-8pt}
\end{figure*}

\subsection{Receptive field at the target resolution}
\label{sec:method:rf}

A DiT layer applies joint attention to the text and image tokens. For a query
$t$ with key set $\mathcal K$,
\begin{equation}
o_t=\sum_{j\in\mathcal K}\frac{e^{\ell_{tj}}}{\sum_{i\in\mathcal K}e^{\ell_{ti}}}\,v_j,
\qquad
\ell_{tj}=\frac{\langle R(p_t)\,q_t,\;R(p_j)\,k_j\rangle}{\sqrt d},
\label{eq:attn}
\end{equation}
where $q_t$ and $k_j$, taken after QK normalization and before RoPE, are the
query and key of one head, $v_j$ is the value, $d$ is the head dimension,
$p$ is a token position and $R(p)$ is the rotary position
embedding~\cite{rope}. Let the image have $s_0\times s_0$ tokens at the native
resolution and $s\times s$ at the target resolution, with $\alpha=s/s_0$,
\eg, $\alpha=4$ at $4096^2$, so that a target position $p$ corresponds to the
native position $p/\alpha$. We call the $\alpha\times\alpha$ target tokens of
one native token a \emph{cell}. In training, every image query attends to the text and the whole image, with one key
per native token. We describe its receptive field, the set of keys it attends to, by its \emph{extent}, the image region it covers, and its \emph{granularity}, the image area each key represents. In training, it has the full extent and the native
granularity, and each image region enters the softmax once per native token it
contains. Without further training, every layer after the attention still
expects the outputs it received in training. We therefore require an
operator at the target resolution to distribute attention over the text and the
image regions as native attention does.

At the target resolution, dense attention and window attention both attend to
target tokens, at a granularity $\alpha^2$ times finer than in training. Dense attention keeps the extent, but each cell now enters the
softmax as $\alpha^2$ keys with nearly identical logits, whose total weight is
about $\alpha^2$ times that of the native key they replace. The image therefore outweighs the text by an extra factor of $\alpha^2$. The cost also grows quadratically with the number of tokens. Window attention attends to
as many keys as in training, but only in a window around the query. The extent shrinks to the window, and objects repeat
across independently generated windows.

Fig.~\ref{fig:analysis} measures these effects on \dev{}, with native
attention as the exact reference. Each operator receives a $4096^2$ input that copies every native token to its cell, so every change comes from the operator. Under every dense-attention operator except
I-Max, the text attention share falls from about 16\% to 2--4\%, and attention shifts toward regions at least 16 native tokens away. Window attention
loses the attention on these far regions and still lowers the text attention share by about 6 percentage points.

To distribute attention as native attention does, an operator must keep the full extent and count each
image region once per native token it
contains, as in training. It must also bound the number of keys. \ours{} meets the first and third requirements with a
distance-dependent key set (Sec.~\ref{sec:method:keys}) and the second with key
weights from native area (Sec.~\ref{sec:method:mass}).

\subsection{Distance-dependent key set}
\label{sec:method:keys}

We follow the common cascade~\cite{hiflow,hrdit,scalediff}. In the first stage, the model generates an image at the native resolution. In the second stage, its latent is upsampled, re-noised~\cite{sdedit} and denoised at the target resolution
with \ours{} in every attention layer and step (Fig.~\ref{fig:method}a). \ours{} groups the image queries into
square query blocks, and the queries of a block share one key set
$\mathcal K_t$ with four parts (Fig.~\ref{fig:method}b), which a kernel can process as one dense unit. The \emph{near field}
$\mathcal W$ consists of the query block and its neighbouring blocks, a window shifted inward at image borders, and keeps its target tokens.

A query also depends on the detail of a few distant regions. The \emph{routed blocks} are a fixed number of other blocks, also kept at the target
resolution, which the first stage selects. One native-resolution forward pass
on the final first-stage latent, re-noised to the starting noise level of the
second stage, records the attention of every layer and head. Each query block
then keeps the blocks outside its near field that receive the most native
attention. The selection is made once per image, layer, head and CFG
branch and reused at all target-resolution steps, so the first stage passes on not only
the layout but also where each query should look.

All other regions mainly provide context. Every remaining cell forms
the \emph{far field} and is pooled into one key and one value by averaging,
\begin{equation}
\bar k_c=\frac{1}{\alpha^2}\sum_{i\in c}k_i,\qquad
\bar v_c=\frac{1}{\alpha^2}\sum_{i\in c}v_i,
\label{eq:pool}
\end{equation}
placed at the native position $\bar p_c$ of the cell centre. Finally, all
\emph{text tokens} are added. The key set covers the whole image without overlap, and its size is independent of the resolution, since it
holds a fixed number of near-field and routed tokens, at most $s_0^2$ pooled keys and the text. Text queries attend to all text tokens and all pooled keys.

Near-field logits use target positions for both the query and the key, so
neighbouring tokens keep distinct offsets at the target scale. All other logits
use native positions, with routed tokens at $p_j/\alpha$, pooled keys at
$\bar p_c$ and the query at $p_t/\alpha$. For routed and pooled keys, the query position is shifted so that the offset counts the distance to the window edge
at the target scale and the distance beyond it at the native scale. This keeps the offsets continuous across the edge of the near field. We write $\ell_{tj}$ for every logit, computed as in Eq.~\eqref{eq:attn} with
the positions that apply to key $j$.

\subsection{Key weights from native area}
\label{sec:method:mass}

The keys of this set represent image areas of different sizes. We say that a key with multiplier $m_j$ on its softmax weight counts $m_j$ times. In training, every key counts once. \ours{} gives
every key the count of the native area it represents
(Fig.~\ref{fig:method}c). A
pooled key and a text token keep $m_j=1$. A target token covers
$1/\alpha^2$ of a native token and gets $m_j=\alpha^{-2}$, which equals a logit
shift of $-\log\alpha^2$. Since text tokens do not change with the resolution,
the text keeps its native weight relative to the image (Fig.~\ref{fig:analysis}).
The near field follows the same rule as a group. Weighting its tokens one by one would not achieve this, because their logits use target positions and do not sum to the weight that native attention gives to the same cells. Inside the window, the queries
keep the exact target-resolution distribution
$\bar o_t=\sum_{j\in\mathcal W}\mathrm{softmax}_j(\ell_{tj})\,v_j$. The window
as a whole, however, receives the total weight $e^{L_t}$ that native attention assigns to its cells, pooled as in Eq.~\eqref{eq:pool},
\begin{equation}
L_t=\operatorname*{LSE}_{c\in\mathcal C(\mathcal W)}
\frac{\langle R(p_t/\alpha)\,q_t,\;R(\bar p_c)\,\bar k_c\rangle}{\sqrt d},
\label{eq:mass}
\end{equation}
where $\operatorname{LSE}$ is the log-sum-exp over the window's cells
$\mathcal C(\mathcal W)$, with the query at its unshifted native position. These pooled cells only set $L_t$ and are not keys themselves. All keys of a query then share one softmax,
\begin{equation}
o_t=\frac{e^{L_t}\,\bar o_t+\sum_{j\in\mathcal K_t\setminus\mathcal W}m_j\,e^{\ell_{tj}}\,v_j}
{e^{L_t}+\sum_{j\in\mathcal K_t\setminus\mathcal W}m_j\,e^{\ell_{tj}}},
\label{eq:nva}
\end{equation}
where $j$ runs over the routed tokens, pooled keys and text tokens, with
$(k_j,v_j,p_j)=(\bar k_c,\bar v_c,\bar p_c)$ for a pooled key. Equivalently, every
near-field logit is shifted by $L_t-\operatorname{LSE}_{i\in\mathcal W}\ell_{ti}$,
so Eq.~\eqref{eq:nva} is a single softmax over all keys. Each image region is
thus counted once per native token it contains, as in training.

At $\alpha=1$, every cell is a single token, all $m_j$ are 1, the positions and
their shifts coincide with those of Eq.~\eqref{eq:attn}, and $L_t$ equals the
log-sum-exp of the near-field logits, so Eq.~\eqref{eq:nva} reduces to
Eq.~\eqref{eq:attn} and \ours{} equals the pretrained attention by
construction. Because the key set is bounded, the attention cost is linear in the number of tokens for a fixed native resolution and text length, whereas dense attention is quadratic. A block-sparse kernel realizes this cost.

\section{Experiments}
\label{sec:exp}

\begin{figure*}[t]
\centering
\includegraphics[width=\textwidth]{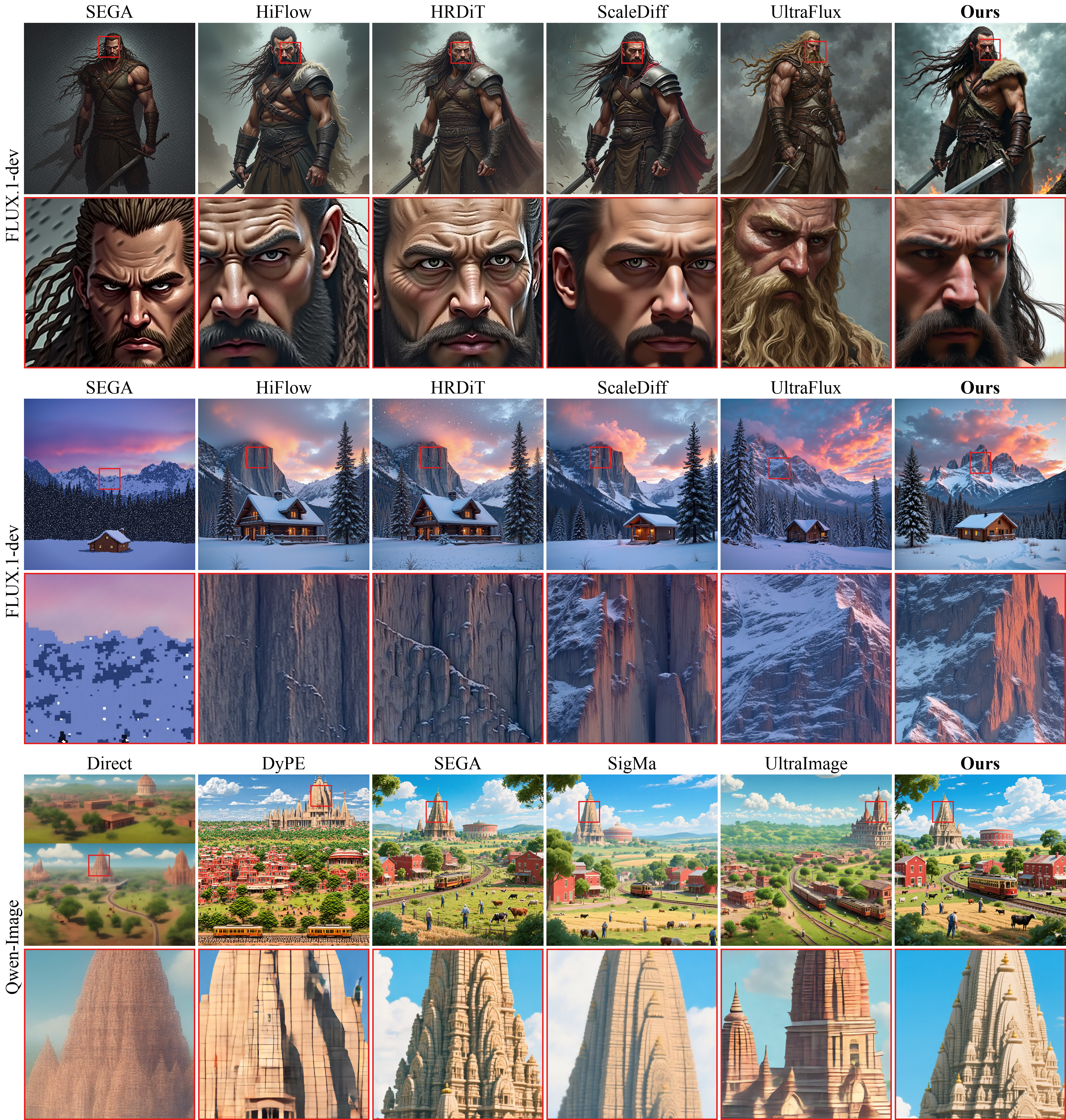}
\caption{\textbf{Qualitative comparison at $4096^2$} on \dev{} (top two
blocks) and \qwen{} (bottom). In each block, the upper row shows whole images,
with red boxes marking the same object across layouts that differ between
methods. The lower row shows the boxed regions as $512^2$ crops at full resolution. Best viewed zoomed in.}
\label{fig:qual}
\vspace{-8pt}
\end{figure*}

\begin{table*}[t]
\centering
\caption{Comparison on \afk{} (195 prompts) at $4096^2$. Each method runs its full pipeline with its official settings under the same evaluation protocol. HPS is
$\times100$. \textbf{Bold} and \underline{underline} mark the best and second-best training-free
results. s/img is the end-to-end latency on one A800 GPU. Attention is the attention pattern that a method uses at $4096^2$.}
\label{tab:main}
\resizebox{0.85\textwidth}{!}{%
\begin{tabular}{lllcccccccc}
\toprule
Method & Attention & Venue & FID\down & pFID\down & IS\up & pIS\up & CLIP\up & HPS\up & MUSIQ\up & s/img\down\\
\midrule
\multicolumn{11}{c}{\textbf{(a) \dev{}}}\\
\midrule
\multicolumn{11}{l}{\emph{Fine-tuned at $4096^2$}}\\
Diffusion-4K~\cite{diffusion4k} & Dense & CVPR'25 & 146.1 & 63.4 & 6.81 & 5.23 & 27.64 & 24.66 & 33.7 & 146\\
UltraFlux~\cite{ultraflux} & Dense & CVPR'26 & 138.5 & 50.8 & 6.97 & 9.22 & 29.20 & 30.05 & 52.3 & 146\\
\midrule
\multicolumn{11}{l}{\emph{Training-free}}\\
Direct $4096^2$ & Dense & -- & 353.0 & 298.5 & 1.26 & 1.85 & 14.47 & 4.41 & 41.2 & 654\\
DyPE~\cite{dype} & Dense & ICML'26 & 206.1 & 185.6 & 5.40 & 6.81 & 24.67 & 20.38 & 49.2 & 645\\
SEGA~\cite{sega} & Dense & NeurIPS'26 & 169.0 & 124.5 & 6.84 & 7.35 & 26.66 & 24.05 & 49.8 & 644\\
SigMa~\cite{sigma} & Dense & ICML'26 & 260.7 & 269.7 & 4.10 & 5.45 & 20.78 & 14.90 & 47.0 & 671\\
UltraImage~\cite{ultraimage} & Dense & arXiv'25 & 138.6 & 68.9 & 6.72 & 7.06 & 28.82 & 30.61 & 41.4 & 787\\
HiFlow~\cite{hiflow} & Dense & NeurIPS'25 & 137.5 & \textbf{54.6} & \underline{7.09} & 8.50 & 29.10 & 30.74 & 51.3 & 375\\
I-Max~\cite{imax} & Dense & arXiv'24 & 138.1 & 62.6 & \underline{7.09} & 7.11 & 28.89 & 30.67 & 40.9 & 868\\
HRDiT~\cite{hrdit} & Window & ECCV'26 & \underline{136.2} & 64.3 & 6.73 & 9.16 & 29.15 & \underline{30.84} & 53.9 & \underline{213}\\
ScaleDiff~\cite{scalediff} & Window & NeurIPS'25 & 138.1 & 66.1 & 6.86 & \textbf{9.93} & \underline{29.23} & 30.80 & \underline{61.9} & 272\\
\textbf{\ours{} (ours)} & \textbf{\ours{}} & -- & \textbf{133.9} & \underline{57.2} & \textbf{7.72} & \underline{9.38} & \textbf{29.28} & \textbf{31.43} & \textbf{62.3} & \textbf{182}\\
\midrule
\multicolumn{11}{c}{\textbf{(b) \qwen{}}}\\
\midrule
Direct $4096^2$ & Dense & -- & 161.5 & 71.4 & 6.87 & 4.21 & 27.51 & 21.11 & 37.3 & 2413\\
DyPE~\cite{dype} & Dense & ICML'26 & 150.6 & 72.6 & 6.78 & 7.80 & 28.60 & 25.49 & 44.8 & 1951\\
SEGA~\cite{sega} & Dense & NeurIPS'26 & \underline{137.8} & 58.8 & 6.88 & \underline{9.31} & \underline{30.12} & \underline{32.30} & \underline{51.1} & 2018\\
SigMa~\cite{sigma} & Dense & ICML'26 & 142.1 & 54.8 & \underline{7.20} & 6.22 & 29.49 & 29.22 & 42.7 & \underline{1905}\\
UltraImage~\cite{ultraimage} & Dense & arXiv'25 & 140.6 & \textbf{50.8} & 6.69 & 6.50 & 29.35 & 27.73 & 45.7 & 2512\\
\textbf{\ours{} (ours)} & \textbf{\ours{}} & -- & \textbf{134.4} & \underline{53.8} & \textbf{7.25} & \textbf{9.87} & \textbf{30.38} & \textbf{33.13} & \textbf{57.8} & \textbf{561}\\
\bottomrule
\end{tabular}}

\vspace{-4pt}
\end{table*}

\subsection{Setup}
\label{sec:exp:setup}

\paragraph{Implementation details.}
We evaluate \ours{} on \dev{}~\cite{flux} and \qwen{}~\cite{qwenimage}, both
trained at about $1024^2$. The pipeline first generates a $1024^2$ image with
the base model, re-noises its upsampled latent to $\sigma^*=0.89$, and then
denoises at the target resolution with \ours{} in all attention layers. The two stages use 28 and 20
steps with guidance 4.5 on \dev{}, and 40 and 29 steps with CFG~\cite{cfg} 6 and 4 on
\qwen{}. Query blocks have $8\times8$ tokens, with a $3\times3$-block near
field and 42 routed blocks. At $4096^2$, each image query thus attends to 7{,}156 image keys instead of the 65{,}536 of dense attention, plus the text. %
\ours{} runs as a block-sparse Triton~\cite{triton} kernel. All timings are on a single A800 GPU.

\paragraph{Baselines.}
We compare against training-free methods in three categories. The first two use dense attention, with position extrapolation (DyPE~\cite{dype}, SEGA~\cite{sega}, SigMa~\cite{sigma}, UltraImage~\cite{ultraimage}) or with a cascade and guidance (HiFlow~\cite{hiflow}, I-Max~\cite{imax}). The third combines a cascade with window attention (HRDiT~\cite{hrdit}, ScaleDiff~\cite{scalediff}). The last two categories run
on \dev{} only, as in their official code. We also
include direct $4096^2$ sampling and, for reference, the fine-tuned Diffusion-4K~\cite{diffusion4k} and UltraFlux~\cite{ultraflux}.

\paragraph{Evaluation.}
We use the prompts and reference images of Aesthetic-Eval@4096 (\afk{}, 195
pairs)~\cite{diffusion4k} on both models.
On \dev{}, we add LAION-1K, with 1{,}000 captions and 10{,}000 disjoint reference
images from the high-resolution subset of Re-LAION~\cite{laion5b}, which give a
more stable distribution comparison than 195 samples. On it, we compare four baselines, two position-extrapolation methods (DyPE, SEGA) and
two cascades (HiFlow, HRDiT).
Each method generates one image per prompt with a shared seed.
We report FID~\cite{fid} following Diffusion-4K~\cite{diffusion4k} and IS~\cite{inceptionscore} on whole images, and pFID, pKID~\cite{kid} and pIS on full-resolution crops for fine detail, computed with torch-fidelity~\cite{torchfidelity}. On \afk{}, these are 20 crops of $299^2$ per image at identical coordinates in generated and reference images. CLIP
score~\cite{clip,clipscore} measures text alignment, HPS~\cite{hpsv2} human preference, and MUSIQ~\cite{musiq} no-reference quality on full-resolution crops. All numbers, including those of the baselines, are from our runs.

\subsection{Main Results}
\label{sec:exp:main}

\begin{table}[t]
\centering
\caption{\dev{} on LAION-1K at $4096^2$ (1{,}000 captions, 10K
references). pKID is $\times10^3$. Bold: best.}
\label{tab:laion}
\footnotesize
\setlength{\tabcolsep}{2.4pt}
\begin{tabular}{@{}llccccc@{}}
\toprule
Method & Venue & FID\down & pFID\down & pKID\down & CLIP\up & MUSIQ\up\\
\midrule
DyPE~\cite{dype} & ICML'26 & 164.7 & 177.2 & 110.61 & 19.27 & 43.7\\
SEGA~\cite{sega} & NeurIPS'26 & 92.8 & 103.2 & 51.63 & 22.77 & 46.7\\
HiFlow~\cite{hiflow} & NeurIPS'25 & 47.0 & 39.7 & 10.17 & 26.89 & 43.0\\
HRDiT~\cite{hrdit} & ECCV'26 & 47.7 & 40.4 & 10.52 & 26.89 & 41.6\\
\midrule
\textbf{\ours{} (ours)} & -- & \textbf{46.3} & \textbf{38.7} & \textbf{8.57} & \textbf{27.53} & \textbf{57.6}\\
\bottomrule
\end{tabular}

\vspace{-8pt}
\end{table}

Table~\ref{tab:main} compares $4096^2$ generation on \afk{}. On \dev{}, all seven dense-attention pipelines need at least 375\,s per image and reach at most 51.3 in MUSIQ. Among them, direct sampling, DyPE, SEGA and SigMa also break global structure (FID $\ge 169.0$). The two window-attention cascades, HRDiT and ScaleDiff, are faster and sharper but trail \ours{} in FID, IS, CLIP and HPS. \ours{} is
second only to HiFlow in pFID and to ScaleDiff in pIS and obtains
the best FID, IS, CLIP, HPS and MUSIQ among training-free methods on \dev{}. On
\qwen{}, where extrapolation baselines led by SEGA remain strong, \ours{} leads
on FID, IS, pIS, CLIP, HPS and MUSIQ. Without training at $4096^2$, \ours{} also exceeds
the fine-tuned Diffusion-4K on all seven quality metrics and UltraFlux on all
except pFID.

On LAION-1K (Tab.~\ref{tab:laion}), the larger set confirms this picture, as DyPE
and SEGA break the distribution, while \ours{} is best on all five metrics. It
surpasses the strongest cascade, HiFlow, on FID, pFID and pKID and leads MUSIQ by 10.9 over the next best of the four baselines, SEGA.

Fig.~\ref{fig:qual} shows the trade-off behind these numbers. Direct sampling and position extrapolation break global layout or texture. Direct splits the \qwen{} image
into two stacked scenes, DyPE turns the temple tower into a grid of flat
blocks, and on \dev{} SEGA covers the face with diagonal dashes and reduces the
mountain to a pixel mosaic. Cascades keep the native $1024^2$ layout but
distort local detail, turning wrinkles into hard engraved lines (HiFlow, HRDiT)
and rock into vertical streaks (HiFlow) or repeated grooves (HRDiT). \ours{}
keeps both layout and natural local detail, from skin and beard to sunlit
rock and temple tiers.

\ours{} is also the fastest training-free method, with 182\,s per image on \dev{}
and 561\,s on \qwen{}, $1.2\times$ and $3.4\times$ faster than the next fastest, HRDiT (213\,s)
and SigMa (1905\,s). Fig.~\ref{fig:eff} analyzes scaling with resolution.

\begin{table}[t]
\centering
\caption{Ablations on \dev{} at $4096^2$ (\afk{}, 195 prompts). Each row
removes or changes one component of the full method. HPS is $\times100$.
Text queries attend to the image only through the far field, so both variants without it also cut them off from the image.}
\label{tab:ablation}
\footnotesize
\setlength{\tabcolsep}{2.4pt}
\begin{tabular}{@{}lccccccc@{}}
\toprule
Variant & FID\down & pFID\down & IS\up & pIS\up & CLIP\up & HPS\up & MUSIQ\up\\
\midrule
Full \ours{} & 133.9 & 57.2 & 7.72 & 9.38 & 29.28 & 31.43 & 62.3\\
\midrule
w/o far field & 134.4 & 60.4 & 7.55 & 8.45 & 29.05 & 30.82 & 59.6\\
w/o routed blocks & 137.1 & 58.5 & 7.33 & 7.60 & 29.13 & 29.85 & 58.1\\
Near field only & 223.4 & 159.8 & 4.27 & 7.59 & 25.68 & 21.17 & 61.7\\
w/o $L_t$ & 135.3 & 77.5 & 7.63 & 6.84 & 29.17 & 30.50 & 59.5\\
Target-area weights & 135.8 & 59.5 & 7.57 & 8.87 & 29.08 & 31.39 & 62.7\\
\bottomrule
\end{tabular}

\vspace{-8pt}
\end{table}

\subsection{Ablations and Analysis}
\label{sec:exp:abl}

\begin{figure*}[t]
\centering
\includegraphics[width=\textwidth]{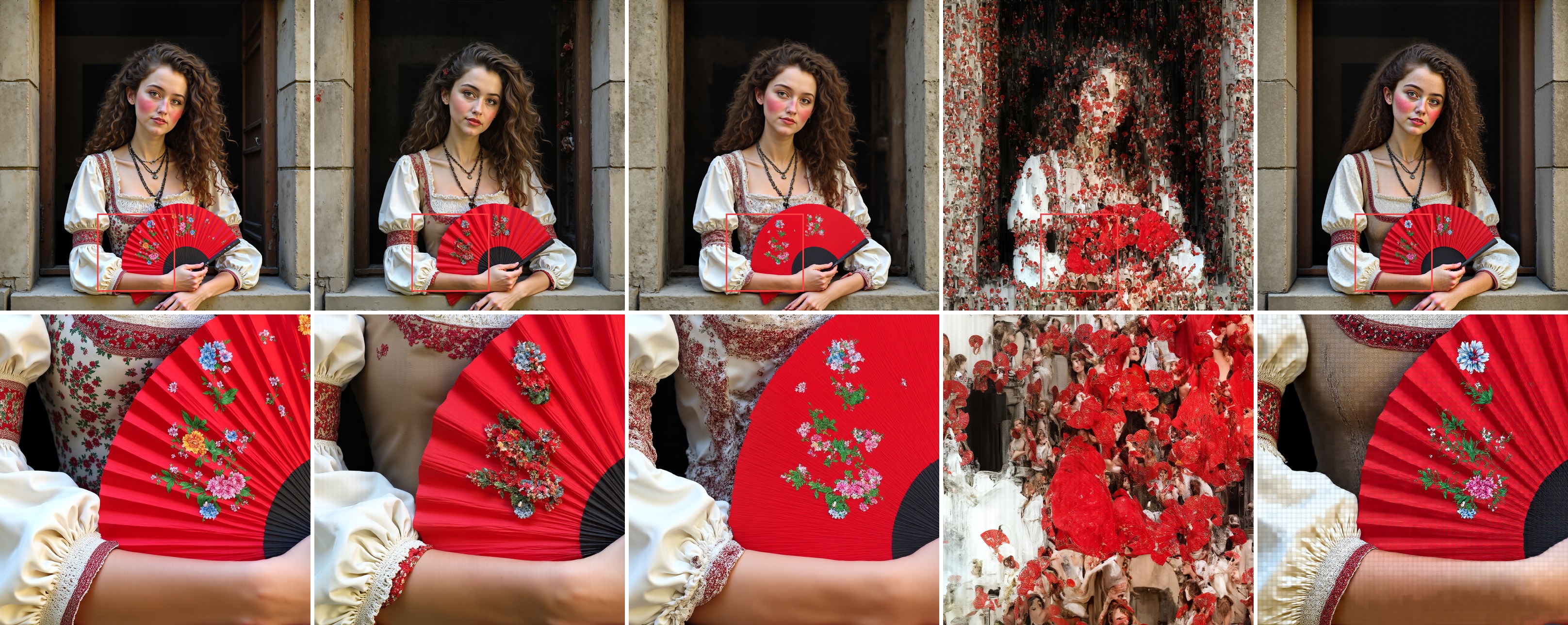}\\
\makebox[0.2\textwidth]{\small Full \ours{}}%
\makebox[0.2\textwidth]{\small w/o far field}%
\makebox[0.2\textwidth]{\small w/o routed blocks}%
\makebox[0.2\textwidth]{\small Near field only}%
\makebox[0.2\textwidth]{\small w/o $L_t$}
\caption{Ablations on one \dev{} sample at $4096^2$. Top: whole image. Bottom:
the red $1024^2$ crop at full resolution.}
\label{fig:ablation}
\vspace{-3pt}
\end{figure*}

Table~\ref{tab:ablation} modifies one component of \ours{} at a time on \dev{} and \afk{} with fixed seeds and the same first stage. Fig.~\ref{fig:ablation} shows one sample.

\paragraph{Near field, routed blocks and far field.}
With only the $3\times3$-block near field and text, queries denoise each window
like an independent native image, forming a mosaic of repeated small red fans
and figures (Fig.~\ref{fig:ablation}). FID rises from 133.9 to 223.4 and HPS
drops from 31.43 to 21.17. MUSIQ stays at 61.7 because it scores local
sharpness, not global coherence. Either distant key set prevents this mosaic,
but the two play distinct roles. Without the 42 routed blocks, queries see
distant regions only as blurry pooled cell averages, so the fan flowers
scatter into loose fragments, lowering HPS to 29.85 and MUSIQ from 62.3 to
58.1. Without the pooled far field, the image stays coherent (HPS 30.82), but
dropping the unrouted cells, each carrying the softmax weight of the
$\alpha^2$ tokens it summarizes, erases the blouse print, raising pFID from 57.2 to 60.4. It also leaves text queries without image keys. Routed blocks thus carry the
most relevant distant content, while the far field keeps the
whole image in view and mainly affects fine detail.

\begin{figure}[t]
\centering
\includegraphics[width=\linewidth]{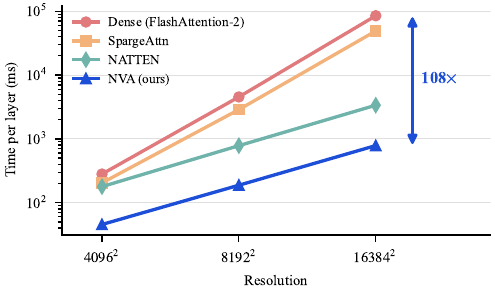}
\caption{Time of one attention layer at each resolution on one A800 GPU
(\dev{} layer 0, median of 9 calls), on log axes.}
\label{fig:eff}
\vspace{-8pt}
\end{figure}

\begin{figure}[t]
\centering
\includegraphics[width=\linewidth]{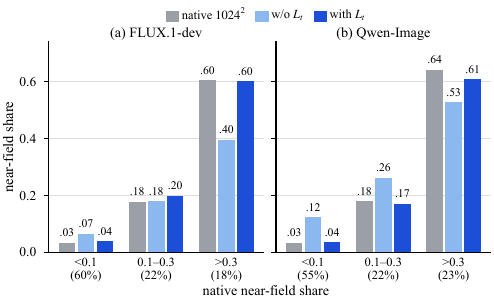}
\caption{Share of image attention that falls in the $3\times3$-block near field,
at the first $4096^2$ step, compared with native $1024^2$ attention of the same
query block. Mean over (layer, head, query block) units grouped by
native near-field share (percent of units in parentheses),
with 24 prompts on \dev{} and 10 on \qwen{} (conditional branch) and the key weights of the main comparison.}
\label{fig:analysis_near_share}
\vspace{-8pt}
\end{figure}

\paragraph{$L_t$ and key weights.}
Without $L_t$ (Eq.~\eqref{eq:mass}), each near-field
token is weighted individually by $\alpha^{-2}$, like a routed token, instead
of the window receiving the total weight that native attention assigns to its cells.
A regular block
grid appears on the sleeve, skin and background (Fig.~\ref{fig:ablation}),
causing the largest detail loss among single-component variants, with pFID rising from 57.2 to 77.5 and pIS dropping from 9.38 to 6.84. Without $L_t$, the near field receives too little attention where native attention is local and too much where it is not, and $L_t$ restores the native share in both cases (Fig.~\ref{fig:analysis_near_share}). Weighting keys by the target area they
cover instead, with each target token counting as one and each pooled key as
$\alpha^2$, lowers the text attention share in proportion to each
target-resolution query's smaller area. On \dev{}, it leaves
HPS nearly unchanged at 31.39 while raising MUSIQ to 62.7 but worsening pFID to 59.5.
Across all variants except near-field-only, whole-image FID varies by at most 3.2 (133.9--137.1), as downsampling
removes the high-resolution detail that patch metrics capture.

\paragraph{Efficiency.}
Fig.~\ref{fig:eff} times one \dev{} attention layer on
an A800 GPU across $4096^2$--$16384^2$, including the pooling, routing and merging steps of \ours{}.
Each doubling of the side length multiplies the time of \ours{} by $4.1$--$4.2$, close to the factor of 4 of a linear cost. For dense FlashAttention-2~\cite{flashattention2}, the factor is $16$--$19$. \ours{} is thus $6.2\times$, $24\times$ and $108\times$ faster at $4096^2$, $8192^2$ and $16384^2$ ($0.79$\,s vs.\ $85.0$\,s at $16384^2$). SpargeAttn~\cite{spargeattn} still grows by $14$--$17\times$ per doubling and is $63\times$ slower than \ours{} at $16384^2$ ($49.1$\,s vs.\ $0.79$\,s). NATTEN~\cite{natten} ($80\times80$ window) is also linear, but
\ours{} is $3.9$--$4.3\times$ faster at every resolution while attending to a
comparable number of keys ($6{,}400$ vs.\ $7{,}156$ image keys), because all 64 queries of a block share one key set.
End to end, the first stage, the non-attention layers and VAE decoding are unchanged. They limit the
speedup over the same cascade with dense attention to $2.5\times$ and
$2.7\times$ at $4096^2$ on \dev{} and \qwen{} and to $8.4\times$ at $8192^2$ on \dev{}. The gap widens with the resolution, because only dense attention has a quadratic cost.

\section{Conclusion}
\label{sec:conclusion}
We presented \ours{}, a training-free attention operator for sampling a pretrained DiT beyond its native resolution. Dense attention counts every image area many times, and window attention does not count the areas outside its window. \ours{} counts every area once with a bounded number of keys. It pools distant image areas and keeps target tokens in the near field and routed blocks, with every key weighted by the native area it covers. On \dev{} and \qwen{} at $4096^2$, \ours{}
leads all training-free methods on FID, IS, CLIP, HPS and MUSIQ, and its
attention cost is linear in the number of tokens. These results suggest
that a pretrained model can generate far beyond its native resolution if each
query attends to the image as in training.

\FloatBarrier
{
    \small
    \bibliographystyle{ieeenat_fullname}
    \bibliography{main}
}

\end{document}